\documentclass[conference]{IEEEtran}
\IEEEoverridecommandlockouts
\usepackage{cite}
\usepackage{url}
\usepackage{amsmath,amssymb,amsfonts}
\usepackage{algorithmic}
\usepackage{graphicx}
\usepackage{textcomp}
\usepackage{xcolor}
\def\BibTeX{{\rm B\kern-.05em{\sc i\kern-.025em b}\kern-.08em
    T\kern-.1667em\lower.7ex\hbox{E}\kern-.125emX}}
\newcommand{\argmin}{\mathop{\rm arg~min}\limits}
\usepackage{nth}
\usepackage[keeplastbox]{flushend}
\usepackage{subcaption}
\usepackage{breqn}
\usepackage{color}
\begin{document}

\title{P$^2$Net: A Post-Processing Network for Refining Semantic Segmentation of LiDAR Point Cloud based on Consistency of Consecutive Frames\\
\thanks{* Corresponds to Weimin Wang.}

\thanks{This paper is based on results obtained from a project commissioned by the New Energy and Industrial Technology Development Organization (NEDO). Hiroshi Ishikawa was partially supported by JSPS KAKENHI Grant number JP20H00615.}
}
\author{\IEEEauthorblockN{Yutaka Momma}
\IEEEauthorblockA{
\textit{Waseda University},\\
\textit{National Institute of Advanced Industrial}\\ 
\textit{Science and Technology},
Tokyo, Japan \\
y.momma.sept11@ruri.waseda.jp}
\and
\IEEEauthorblockN{Weimin Wang$^{*}$}
\IEEEauthorblockA{
\textit{
National Institute of Advanced Industrial}\\ 
\textit{Science and Technology},
Tokyo, Japan \\
weimin.wang@aist.go.jp}
\and
\IEEEauthorblockN{Edgar Simo-Serra}
\IEEEauthorblockA{
\textit{Waseda University}\\
Tokyo, Japan \\
ess@waseda.jp}
\and
\IEEEauthorblockN{Satoshi Iizuka}
\IEEEauthorblockA{
\textit{University of Tsukuba}\\
Tsukuba, Japan \\
iizuka@cs.tsukuba.ac.jp}
\and
\IEEEauthorblockN{Ryosuke Nakamura}
\IEEEauthorblockA{
\textit{
National Institute of Advanced Industrial}\\ 
\textit{Science and Technology},
Tokyo, Japan \\
r.nakamura@aist.go.jp
}
\and
\IEEEauthorblockN{Hiroshi Ishikawa}
\IEEEauthorblockA{
\textit{Waseda University}\\
Tokyo, Japan \\
hfs@waseda.jp}
}

\maketitle

\begin{abstract}
We present a lightweight post-processing method to refine the semantic segmentation  results of point cloud sequences. Most existing methods usually segment frame by frame and encounter the inherent ambiguity of the problem: based on a measurement in a single frame, labels are sometimes difficult to predict even for humans. To remedy this problem, we propose to explicitly train a network to refine these results predicted by an existing segmentation method. The network, which we call the P$^2$Net, learns the consistency constraints between ``coincident'' points from consecutive frames after registration. We evaluate the proposed post-processing method both qualitatively and quantitatively on the SemanticKITTI dataset that consists of real outdoor scenes. The effectiveness of the proposed method is validated by comparing the results predicted by two representative networks with and without the refinement by the post-processing network. Specifically, qualitative visualization validates the key idea that labels of the points that are difficult to predict can be corrected with P$^2$Net. Quantitatively, overall mIoU is improved from 10.5\% to 11.7\% for PointNet \cite{qi2017pointnet} and from 10.8\% to 15.9\% for PointNet++ \cite{qi2017pointnetplusplus}.
\end{abstract}


\section{INTRODUCTION}

Semantic segmentation of point cloud is a classical and important task in fields from computer vision to robotics and autonomous driving.  In particular, the detection of such objects as lanes, pedestrians, and obstacles is a crucial step for the autonomous vehicle or robotic systems. CNN-based methods have made it possible to achieve unprecedented performance for image classification or objection detection in images \cite{matterport_maskrcnn_2017,Kirillov2019PanopticS}. However, it is susceptible to stably acquire regular images with cameras
under harsh conditions in the real environment, such as sunlight overexposure in bright days or low lighting conditions in the darkness of nights, let alone the subsequent prediction.  Point cloud data acquired with LiDAR sensors are much more robust to these conditions. 

\begin{figure}[t!]
    \centering
    \includegraphics[width=.45\textwidth]{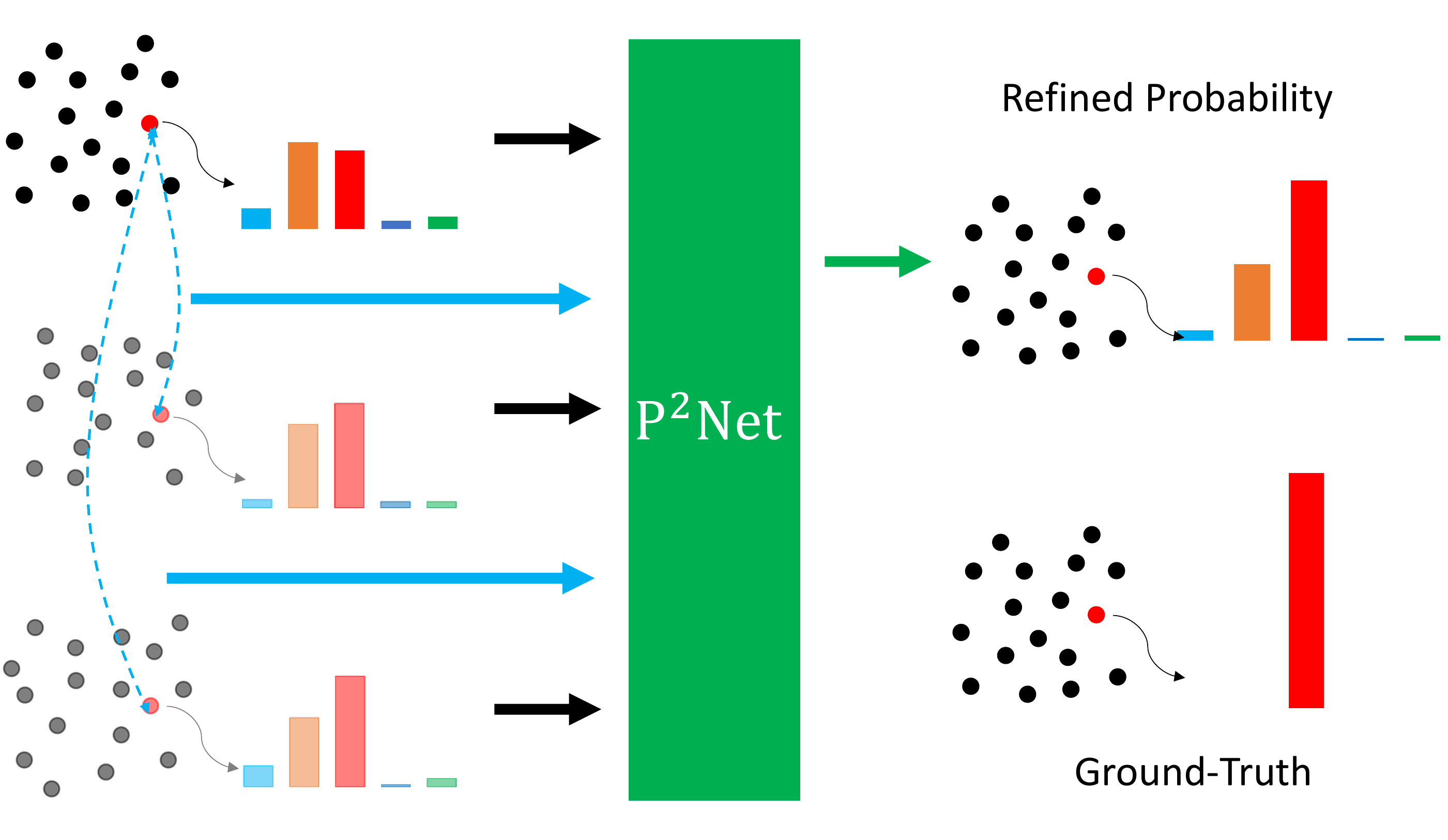}
    \caption{P$^2$Net refines the prediction probabilities of each point in the current frame with the information from nearest neighboring points in previous frames.}
    \label{fig:banner}
\end{figure}
\begin{figure*}[h!]
    \centering
    \includegraphics[width=.97\textwidth]{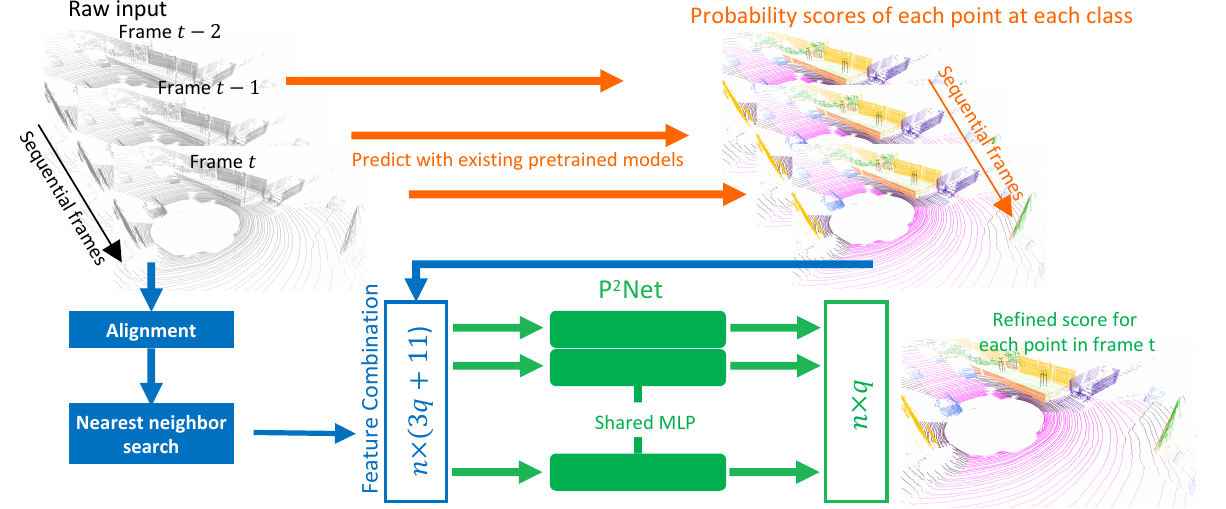}
    \caption{Overview of the proposed refinement scheme. Although the scheme can process any number of sequential frames, we set it to 3 frames in this work. For raw input of 3 frames $t,t-1,t-2$ , we predict the scores of each label for each point with existing pre-trained models, as the orange arrows show. Then, we combine features such as predicted scores and relations with nearest neighbor points in sequential frames $t-1$ and $t-2$. Finally, we feed these combined features to train P$^2$Net to learn how to refine the predicted scores for frame $t$. Here, $n$ is the number of points in frame $t$, $q$ is the number of semantic classes, and 11 indicates the dimension of the combined features.}
    \label{fig:whole}
\end{figure*}

Due to the fundamental representation difference of images and point clouds, it is difficult to directly apply existing CNN-based methods to point clouds. To solve this, previous methods tried to convert 3D point clouds into 2D images. An approach is to capture the image of a 3D visualized point cloud from some viewpoint, for example, Bird's eye view (BEV) images from the top. Generated BEV images are further fed to train networks based on object-detection architectures such as YOLO \cite{yolov3} or Faster-RCNN \cite{ren2015faster}). Another idea introduced by Wu et al. \cite{wu2017squeezeseg,wu2018squeezesegv2} is to convert one scan of the point cloud to the range image by equirectangular projection and apply existing 2D CNN-based methods. Instead of only one viewpoint of the point cloud, Chen et al. \cite{chen2017multi} propose the Multi-view 3D (MV3D) method that fuses the features of BEV, front view, as well as the corresponding RGB images. These methods are usually limited to handle with full 3D point clouds due to the dimensional lack of the expression of the occlusion when projecting 3D data to 2D. 

To keep 3D information as well as make CNN-based methods applicable, Maturana et al. \cite{maturana2015voxnet} propose the voxelization of the point cloud for object detection with 3D CNN method VoxNet. Although a multi-resolution approach is also proposed to improve memory efficiency, there is a trade-off between the voxelized resolution and memory utilization. Instead of compressing multiple points to one voxel, Zhou et al. \cite{yin2017voxelnet} propose to extract the global feature of all points in a voxel with a PointNet-based method \cite{qi2017pointnet} and perform object detection with a region proposal network (RPN) \cite{ren2015faster}.
Voxelization of 3D point cloud can both keep the 3D information of the point cloud and be capable of applying 3D CNN. Nevertheless, the tradeoff between the fineness and the memory consumption has to be made for the voxel-based methods.

Thus, Multilayer perceptron (MLP)-based methods \cite{qi2017pointnet,qi2017pointnetplusplus,qi2019deep} are further proposed to make it able to directly and flexibly process point cloud data. 
Subsequent graph-based methods \cite{Wang2018DynamicGC, Te2018RGCNNRG} are also proposed. Feasibility and promising performance were shown on benchmark datasets at either object-level \cite{Wu20143DSA,shapenet2015} or even scene-level \cite{armeni_cvpr16,Dai2017ScanNetR3,Geiger2013IJRR,behley2019iccv}. Either voxel-based or PointNet-based methods usually process the point cloud data frame-by-frame. To further utilize the temporal information of sequences, Luo et al. \cite{Luo2018FastAF} propose to detect on BEV images sequences. Choy et al.  \cite{mink} propose a 4D sparse convolution Minkowski ConvNets to segment on 4D voxels. To directly process dynamic point clouds,  Liu et al. \cite{Liu2019MeteorNetDL} further propose MeteorNet by extracting features from points grouped across sequences.

With the observation of false segmentation that may come from the occluded points, which can be difficult to tell even for humans, we propose the $P^2$Net to refine the results with spatial consistency that is similar to chained-flow grouping in \cite{Liu2019MeteorNetDL}. The difference from \cite{Liu2019MeteorNetDL} is that we explicitly integrate the probability results predicted by an existing method, as shown in Fig\ref{fig:banner}, for flexibility and efficiency concerns.
Since the inputs of the network are derived from coordinates and probabilities of the points, basically $P^2$Net can be trained to refine the results predicted by any segmentation method. The network is designed to learn how to refine the predicted class probability with prediction results of previous frames as well as other geometric relationships to the nearest neighboring points in other frames. We evaluate the feasibility and efficiency of the proposed method by refining results predicted with PointNet \cite{qi2017pointnet} and PointNet++ \cite{qi2017pointnetplusplus} on SemanticKITTI dataset \cite{behley2019iccv}.

\section{Method}

\subsection{Overview}
There are three steps of the refinement system as shown in  Fig. \ref{fig:whole}.
First, sequential point clouds are input and the classification probability score of each point is predicted with a pre-trained model. Then, nearest neighbors of each point are searched and scored across sequential frames which are aligned with SLAM technologies.
The predicted scores are well as other geometric information of a point and relative differences with its nearest neighbor points are fused as the input for $P^2$Net. Finally, the MLP-based P$^2$Net is trained by feeding these combined inputs to refine the previous probability results.

\subsection{Pre-trained models}
\label{subsec:base_model}
Since the proposed is a post-processing scheme, we assume there exists a pre-trained model such as PointNet \cite{qi2017pointnet} or PointNet++ \cite{qi2017pointnetplusplus}. We denote the set of points in frame $t$ by ${P}^t = \{ \textbf{p}^t_1, \textbf{p}^t_2, \ldots, \textbf{p}^t_n \}$, where $n$ is the number of points, $\textbf{p}^t_i=[x^t_i,y^t_i,z^t_i,r^t_i] \in \mathbb{R}^4$ is the $i$-th point in frame $t$, where $x,y,z$, and $r$ represent the 3D coordinates and the reflected intensity. For the input point cloud ${P}^t$, the pretained model $f$ gives the predicted probability scores for each point as: 
\begin{equation}
    C^t = f(P^t),
    \label{eq:base}
\end{equation}
where $C^t = \{ \textbf{c}^t_1, \textbf{c}^t_2, \ldots, \textbf{c}^t_n \}$ and  $\textbf{c}^t_i=[{c}^t_{i,1},{c}^t_{i,2}, \ldots,{c}^t_{i,q}]$. Here, ${c}^t_{i,j}$ is the score for class $j$ at point $\textbf{p}^t_i$, and $q$ is the number of semantic classes. 

\subsection{Alignment of multiple frames}
\label{subsec:multi_frame}
For frames in ordinary videos, pixels are aligned with a well-defined adjacency relation. Thus, kernels in neural networks can further extract features either along spatial direction or in the temporal direction. That is to say,  the feature of each pixel can be computed via the kernel convolution with its spatially or temporally adjacent pixels. However, there are no such fixed relations between points in the point cloud. In this work, for any point in the current frame, we consider its nearest neighbor as its spatially and temporally adjacent point for feature combination. To establish adjacency relations between different frames by nearest neighbor search, we align the frames into a common coordinate system. The alignment can be achieved with existing Simultaneous Localization and Mapping (SLAM) methods such as LOAM \cite{Zhang2014LOAMLO} and SuMa \cite{Behley2018EfficientSS}. We assume sequential frames have been aligned.

\subsection{Combination of features from multiple frames}
\label{subsec:create}

\subsubsection{Scores of sequential frames}
\label{subsubsec:score}
 As defined in Eq. \ref{eq:base}, the probability scores $C^t=\{ \textbf{c}^t_1, \textbf{c}^t_2, \ldots, \textbf{c}^t_n \}$, $C^{t-1} = \{ \textbf{c}^{t-1}_1, \textbf{c}^{t-1}_2, \ldots, \textbf{c}^{t-1}_m \}$, and $C^{t-2} = \{ \textbf{c}^{t-2}_1, \textbf{c}^{t-2}_2, \ldots, \textbf{c}^{t-2}_l \}$ are predicted for $P^t$, $P^{t-1}$, and $P^{t-2}$. Note that $l$, $m$, and $n$ may differ due to the different numbers of points.

\subsubsection{Nearest neighbor in sequential frames}
As explained above, the current frame $P^t = \{ \textbf{p}^t_1, \textbf{p}^t_2, \ldots, \textbf{p}^t_n \}$ and subsequent frames $P^{t-1} = \{ \textbf{p}^{t-1}_1, \textbf{p}^{t-1}_2, \ldots, \textbf{p}^{t-1}_m \}$, $P^{t-2} = \{ \textbf{p}^{t-2}_1, \textbf{p}^{t-2}_2, \ldots, \textbf{p}^{t-2}_l \}$ have been aligned into a common coordinate system. For $\textbf{p}^t_i$ in frame $t$, its nearest neighbor is denoted by $\textbf{p}^{(t, t-1)}_i$ and $\textbf{p}^{(t, t-2)}_i$ for frame $t-1$ and $t-2$:
\begin{equation}
\begin{split}
\label{eq:nn}
    \textbf{p}^{(t, t-1)}_i & =  \argmin_{\textbf{p}^{t-1}_j\in P^{t-1}} d(\textbf{p}^t_i,\textbf{p}^{t-1}_j)  \\
    \textbf{p}^{(t, t-2)}_i & =  \argmin_{\textbf{p}^{t-2}_k\in P^{t-2}} d(\textbf{p}^t_i,\textbf{p}^{t-2}_k) 
\end{split}
\end{equation}
where $d(\textbf{p}^t_i,\textbf{p}^{t-1}_j)$ and $d(\textbf{p}^t_i,\textbf{p}^{t-2}_k)$ are the Euclidean distance from $\textbf{p}_i^t$ to $\textbf{p}_j^{t-1}$ and $\textbf{p}_k^{t-2}$, respectively:
\begin{equation}
\begin{split}
\label{eq:dis_nn}
    d(\textbf{p}^t_i,\textbf{p}^{t-1}_j) & =     {\left \|[x_i^t, y_i^t, z_i^t] - [x_j^{t-1},y_j^{t-1},z_j^{t-1}] \right \|}_2\\
    d(\textbf{p}^t_i,\textbf{p}^{t-2}_k) & =     {\left \|[x_i^t, y_i^t, z_i^t] - [x_k^{t-2},y_k^{t-2},z_k^{t-2}] \right \|}_2
\end{split}
\end{equation}

After finding the nearest neighbor points in the subsequent frames with Eq.~\ref{eq:nn}, we have the nearest neighbor point sets $P^{(t,u)}=\{\textbf{p}^{(t,u)}_1, \textbf{p}^{(t,u)}_2, \ldots, \textbf{p}^{(t,t-1)}_n \}$ of $P^t$ for $u=t-1,t-2$. Similar to the notation of $\textbf{p}_i^t$ in \S \ref{subsec:base_model}, we denote the components of $\textbf{p}_i^{(t,u)}$ by $[x^{(t,u)}_i,y^{(t,u)}_i,z^{(t,u)}_i,r^{(t,u)}_i]$,
consisting of the spatial coordinates and the reflected intensity of the nearest neighboring point in frame $u$.
\subsubsection{Combined features}
Then, we  gather the probability score vectors of the neighboring points as well as the spatial and reflected intensity difference for the combined features. Specifically, we denote the feature set for point cloud $P^t$ by $P^{'t} = \{ \textbf{p}^{'t}_1, \textbf{p}^{'t}_2, \ldots, \textbf{p}^{'t}_n \}$, which is derived from the probability scores $C^t$, $C^{t-1}$, and $C^{t-2}$ generated in \S \ref{subsubsec:score}, and the nearest neighbor point set $P^{(t,t-1)}$and $P^{(t, t-2)}$.
\begin{eqnarray}
    \textbf{p}^{'t}_i = [ \Delta \textbf{p}^{(t,t-1)}_i, d^{(t,t-1)}_i, \textbf{c}^{(t,t-1)}_i, \qquad \qquad \qquad \nonumber \\
    \qquad \qquad \qquad \Delta \textbf{p}^{(t,t-2)}_i, d^{(t,t-2)}_i, \textbf{c}^{(t,t-2)}_i, r^{t}_i, \textbf{c}^{t}_i], \nonumber
\end{eqnarray}
where $\Delta \textbf{p}$ denotes the difference of 3D coordinates and reflected intensity, i.e.,  for $u = t-1$ and $t-2$:
\begin{align*}
    \Delta \textbf{p}^{(t,u)}_i&=[ \Delta x^{(t, u)}_i, \Delta y^{(t, u)}_i, \Delta z^{(t, u)}_i, r^{(t, u)}_i ] \\
     \Delta x^{(t,u)}_i & =  x^{(t,u)}_i - x^{t}_i \\
        \Delta y^{(t,u)}_i & = y^{(t,u)}_i - y^{t}_i \\
        \Delta z^{(t,u)}_i & = z^{(t,u)}_i - z^{t}_i.
\end{align*}

Also, the distance $d^{(t, u)}_i$ with nearest neighbor points is defined as:
\begin{align*}
        d^{(t, u)}_i &= {\left \| [\Delta{x}^{(t, u)}_i,  \Delta{y}^{(t, u)}_i, \Delta{z}^{(t, u)}_i]\right \|}_2
\end{align*}for $u = t-1, t-2$.
All told, $\textbf{p}^{'t}_i$ has $(3q+11)$-dimensions as shown in Fig. \ref{fig:whole}. Thereinto, $\Delta \textbf{p}^{(t,t-1)}_i$ and $\Delta \textbf{p}^{(t,t-2)}_i$ are $four-$dimensions; $ d^{(t,t-1)}_i$, $d^{(t,t-2)}_i$, $r^{t}_i$ are $one-$dimension; $\textbf{c}^{(t,t-1)}_i$, $\textbf{c}^{(t,t-2)}_i$, $\textbf{c}^{t}_i $ are class number $q-$dimensions.

\subsection{Refinement of scores with trainable P$^2$Net}
To keep the correspondence between the input combined features and the output refined score probabilities, we utilize the MLPs as the structure of P$^2$Net, which is similar to PointnNet \cite{qi2017pointnet}. Concretely, the layer sizes are $(3\times q+11, 128,1024,512,256,128,q)$. Batch normlization\cite{ioffe2015batch} is used for all layers except the last layer with ReLU. The combined feature set $P^{'t}$ generated in \S \ref{subsec:create} are fed to P$^2$Net to train the network.

\begin{table*}[h!]
	\centering

	\renewcommand{\arraystretch}{1.4}
    \begin{tabular}{c|c|c|c|c|c|c|c|c|c|c|c}
    	\hline
        Sequence & 00 & 01 & 02 & 03 & 04 & 05 & 06 & 07 & 08 & 09 & 10 \\
		\hline
    	number of frames & 4541 & 1101 & 4661 & 801 & 271 & 2761 & 1101 & 1101 & 4071 & 1591 & 1201 \\
    	\hline
    \end{tabular}
    \caption{The numbers of frames in available sequences of SemanticKITTI dataset. In this work, we use sequences 00, 01, 02, 03, 06, 07, 09, and 10 for training; 08 for validation; and 04 and 05 for testing.}
	\label{tab:frames}
\end{table*}
\begin{table*}[h!]
    \centering
    \renewcommand{\arraystretch}{1.4}
        \begin{tabular}{l|c|ccc}
            \hline
            Method & Dataset & Number of points per frame & Learning rate & epochs \\
            \hline
            PointNet\cite{qi2017pointnet} & SemanticKITTI \cite{behley2019iccv} & 50,000 & $0.01 \times 0.9^{\mathrm{epoch}}$ & 47 \\
            PointNet++\cite{qi2017pointnetplusplus} & SemanticKITTI \cite{behley2019iccv} & 45,000 & $0.003 \times 0.9^{\mathrm{epoch}}$ & 47 \\
            P$^2$Net(trained w/ PointNet) & {P$^{'t}$} with {C$^t$} predicted by PointNet  & 75,000 & $0.01 \times 0.9^{\mathrm{epoch}}$ & 43 \\
            P$^2$Net(trained w/ PointNet++) & {P$^{'t}$} with {C$^t$} predicted by PointNet++ & 75,000 & $0.005 \times 0.9^{\mathrm{epoch}}$ & 25 \\
            \hline
        \end{tabular}
    \caption{Hyper parameters set in this work.}
    \label{tab:train}
\end{table*}
\begin{table*}[h!]
    \centering
    \tabcolsep = 0.12cm
    \renewcommand{\arraystretch}{1.4}
    \begin{tabular}{lc|ccccccccccccccccccc}
        \hline
        \textbf{Approach} & \rotatebox{90}{\textbf{mIoU}} & \rotatebox{90}{\textcolor[rgb]{0.39, 0.59, 0.98}{\textbf{car}}} & \rotatebox{90}{\textcolor[rgb]{0.39, 0.9, 0.98}{\textbf{bicycle}}} & \rotatebox{90}{\textcolor[rgb]{0.12, 0.24, 0.59}{\textbf{motorcycle}}} & \rotatebox{90}{\textcolor[rgb]{0.31, 0.12, 0.71}{\textbf{truck}}} & \rotatebox{90}{\textcolor[rgb]{0.0, 0.0, 1.0}{\textbf{other-vehicle}  }} & \rotatebox{90}{\textcolor[rgb]{1.0, 0.12, 0.12}{\textbf{person}}} & \rotatebox{90}{\textcolor[rgb]{1.0, 0.16, 0.78}{\textbf{bicyclist}}} & \rotatebox{90}{\textcolor[rgb]{0.59, 0.12, 0.35}{\textbf{motorcyclist}  }} & \rotatebox{90}{\textcolor[rgb]{1.0, 0.0, 1.0}{\textbf{road}}} & \rotatebox{90}{\textcolor[rgb]{1.0, 0.59, 1.0}{\textbf{parking}}} & \rotatebox{90}{\textcolor[rgb]{0.29, 0.0, 0.29}{\textbf{sidewalk}}} & \rotatebox{90}{\textcolor[rgb]{0.69, 0.0, 0.29}{\textbf{other-ground } }} & \rotatebox{90}{\textcolor[rgb]{1.0, 0.78, 0.0}{\textbf{building}}} & \rotatebox{90}{\textcolor[rgb]{1.0, 0.47, 0.20}{\textbf{fence}}} & \rotatebox{90}{\textcolor[rgb]{0.0, 0.69, 0.0}{\textbf{vegetation}}} & \rotatebox{90}{\textcolor[rgb]{0.53, 0.24, 0.0}{\textbf{trunk}}} & \rotatebox{90}{\textcolor[rgb]{0.59, 0.94, 0.31}{\textbf{terrain}}} & \rotatebox{90}{\textcolor[rgb]{1.0, 0.94, 0.59}{\textbf{pole}}} & \rotatebox{90}{\textcolor[rgb]{1.0, 0.0, 0.0}{\textbf{traffic-sign}}} \\
        \hline
        PointNet\cite{qi2017pointnet} & 10.5 & 30.7 & 0.2 & 0.0 & \textbf{0.4} & 0.7 & 0.0 & 0.0 & 0.0 & 40.6 & 19.5 & 26.6 & \textbf{1.5} & 28.0 & 15.8 & 16.7 & 1.3 & 24.1 & 1.0 & 3.4 \\
        + P$^2$Net(trained w/ PointNet) & \textbf{11.7} & 34.5 & \textbf{0.3} & 0.0 & 0.0 & 0.8 & 0.0 & \textbf{0.6} & 0.0 & \textbf{43.7} & 20.3 & \textbf{28.9} & 1.0 & \textbf{30.0} & \textbf{19.8} & \textbf{17.3} & \textbf{1.5} & \textbf{29.5} & \textbf{1.1} & 4.4 \\
        + P$^2$Net(trained w/ PointNet++) & 11.1 & \textbf{39.1} & 0.2 & 0.0 & 0.3 & \textbf{1.2} & \textbf{0.1} & 0.0 & 0.0 & 41.8 & \textbf{22.4} & 26.3 & 0.9 & 28.1 & 14.6 & 16.4 & 1.4 & 22.1 & \textbf{1.1} & \textbf{5.2} \\
        \hline
        PointNet++\cite{qi2017pointnetplusplus} & 10.8 & 20.3 & 0.4 & 0.2 & 0.8 & 0.8 & 0.8 & 5.4 & 0.0 & 37.2 & 6.6 & 24.6 & \textbf{1.3} & 37.7 & 17.7 & 20.8 & 4.1 & 26.4 & 4.8 & 4.9 \\
        + P$^2$Net(trained w/ PointNet) & 12.3 & 16.1 & \textbf{1.8} & \textbf{0.5} & 0.9 & 1.0 & \textbf{3.4} & 4.6 & 0.0 & 43.2 & 9.8 & 29.4 & 1.0 & 42.6 & 17.8 & 21.7 & 5.1 & 30.1 & 5.3 & \textbf{10.8} \\
        + P$^2$Net(trained w/ PointNet++) & \textbf{15.9} & \textbf{54.9} & 0.2 & 0.0 & \textbf{1.3} & \textbf{1.6} & 1.7 & \textbf{17.2} & 0.0 & \textbf{48.0} & \textbf{13.4} & \textbf{31.2} & \textbf{1.3} & \textbf{46.0} & \textbf{22.2} & \textbf{23.8} & \textbf{6.2} & \textbf{33.3} & \textbf{7.4} & 7.9 \\
        \hline
    \end{tabular}
    \caption{Segmentation performances on SemanticKITTI Dataset\cite{behley2019iccv}. As described in \ref{sec:dataset}, sequences 00-03, 06-10 are used to train PointNet and PointNet++ as training and validation sets. Quantitative results in this table are evaluated on sequence 04-05 with the metric which conforms with that in \cite{behley2019iccv}.  We can find that either overall mIoU improves after the refinement with P$^2$Net.}
    \vspace{-0.3cm}
    \label{tab:miou}
\end{table*}

\section{Experiments}
\subsection{Dataset}
\label{sec:dataset}
KITTI dataset
\cite{Geiger2013IJRR} is a well-known benchmark for autonomous driving tasks. Although LiDAR data are available,  only 3D bounding boxes of objects such as vehicles, pedestrians and cyclists are annotated. This is far from enough for semantic segmentation. Behley et al. annotated all point clouds of KITTI dataset directly after aligning them with the SLAM system in \cite{Behley2018EfficientSS} and released it as   
the SemanticKITTI open dataset
\cite{behley2019iccv}. There are 19 valid classes and an outlier class, for a total of $q=20$ classes. We train and  evaluate P$^2$Net on SemanticKITTI dataset. The numbers of available frames in sequences from 00 to 10 are listed in Tab. \ref{tab:frames}. Note that although baselines for methods such as PointNet and PoinNet++ are reported in \cite{behley2019iccv}, pre-trained models of these networks are not provided yet. Thus, we first train the two networks for baseline semantic segmentation on this dataset. The sequences are divided into 00-03, 06-07, 09-10 for training, 08 for validation, and 04-05 for testing.

\begin{figure*}[h!]
    \begin{tabular}{cc}
        \hspace{-0.5cm}
        \begin{minipage}[t]{0.25\linewidth}
            \centering
            \includegraphics[keepaspectratio, scale=0.12]{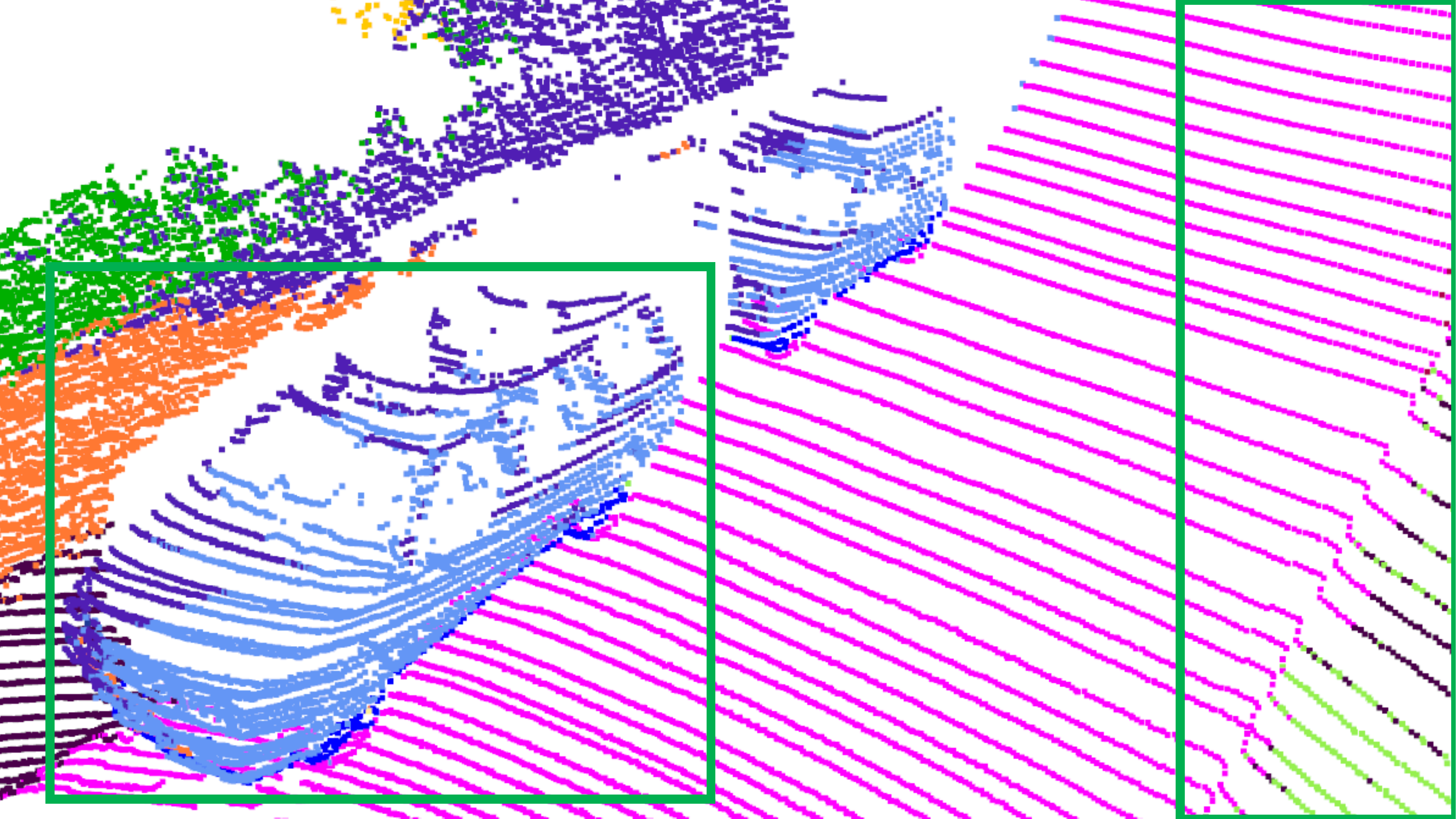}
            \subcaption{}
            \label{fig:pointnet_f1148}
        \end{minipage}
        
        \begin{minipage}[t]{0.25\linewidth}
            \centering
            \includegraphics[keepaspectratio, scale=0.12]{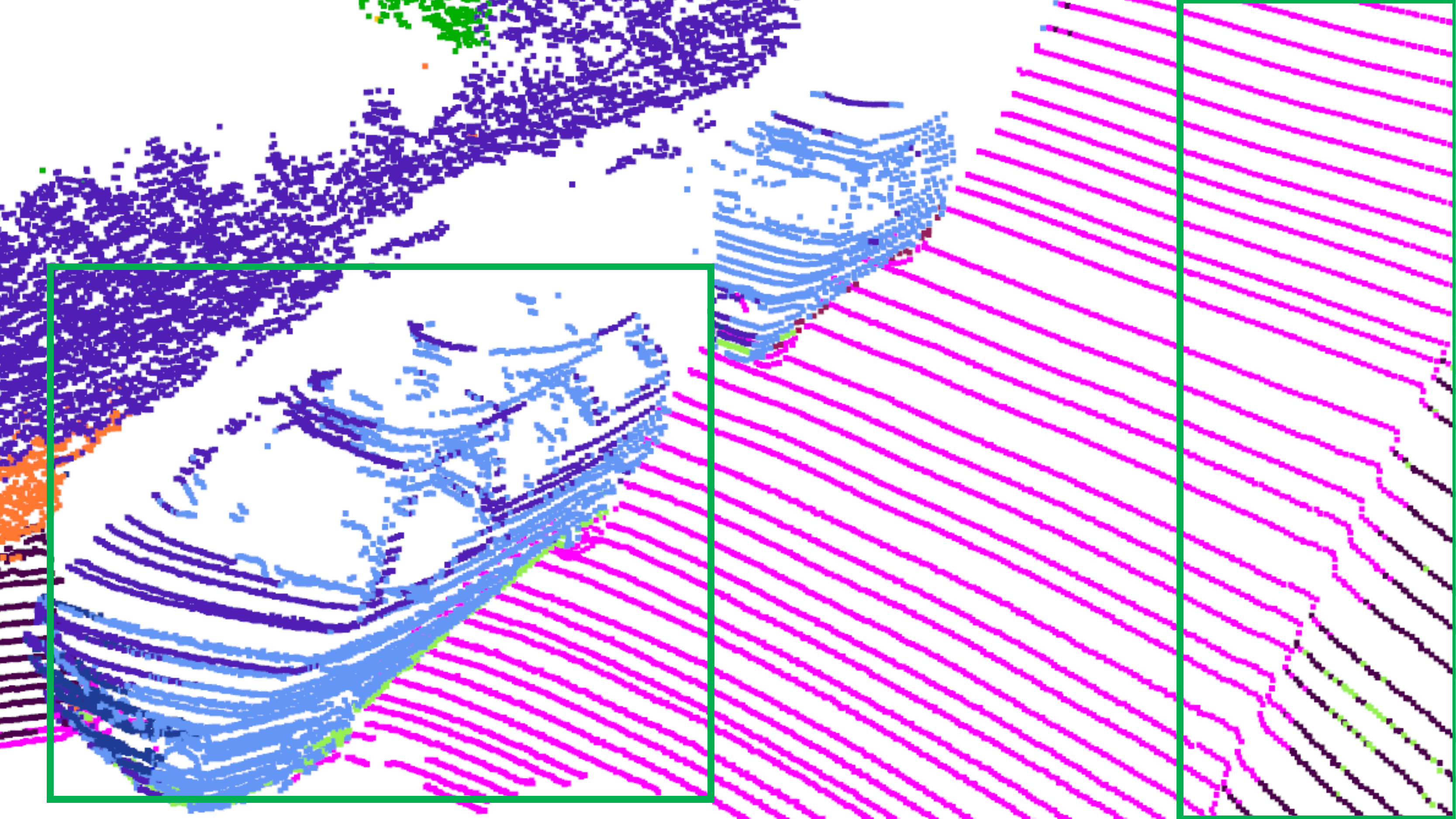}
            \subcaption{}
            \label{fig:pointnet_f1149}
        \end{minipage} 
        
        \begin{minipage}[t]{0.25\linewidth}
            \centering
            \includegraphics[keepaspectratio, scale=0.12]{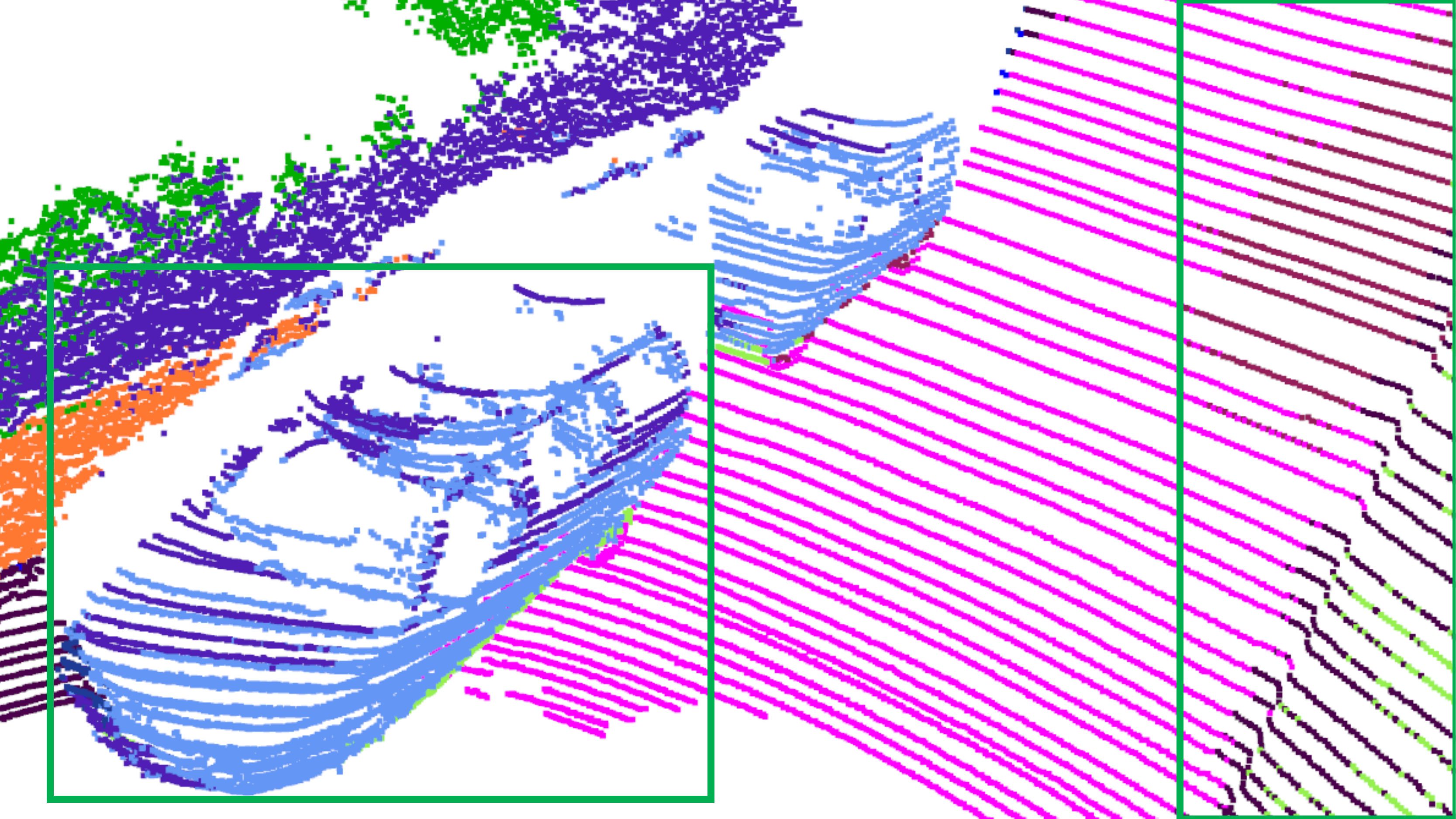}
            \subcaption{}
            \label{fig:pointnet_ali_f1150}
        \end{minipage}
        
        \begin{minipage}[t]{0.25\linewidth}
            \centering
            \includegraphics[keepaspectratio, scale=0.12]{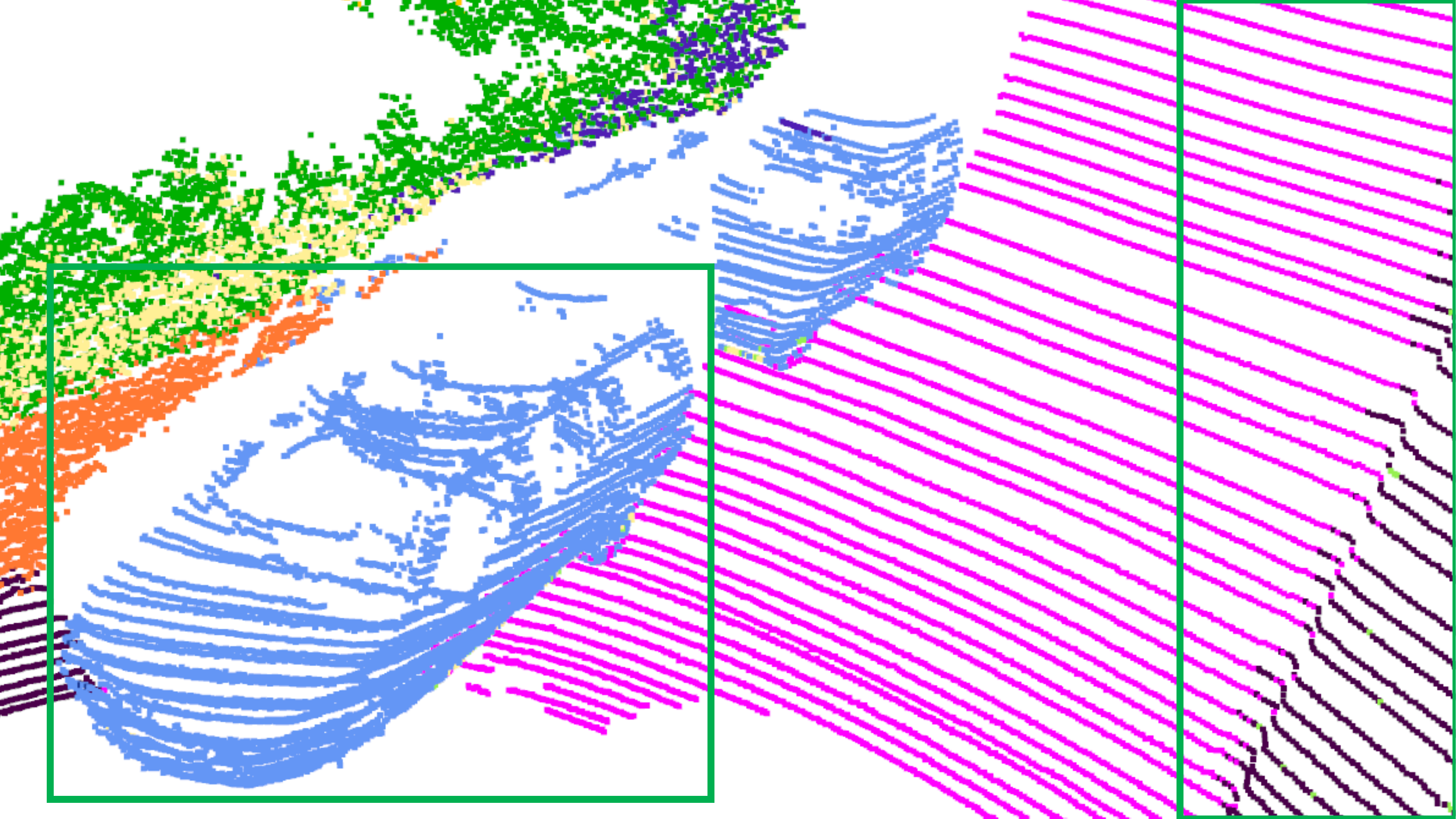}
            \subcaption{}
            \label{fig:pointnet_pp_f1150}
        \end{minipage} \\
        \vspace{-0.2cm}
        \\
        \hspace{-0.5cm}\begin{minipage}[t]{0.25\linewidth}
            \centering
            \includegraphics[keepaspectratio, scale=0.12]{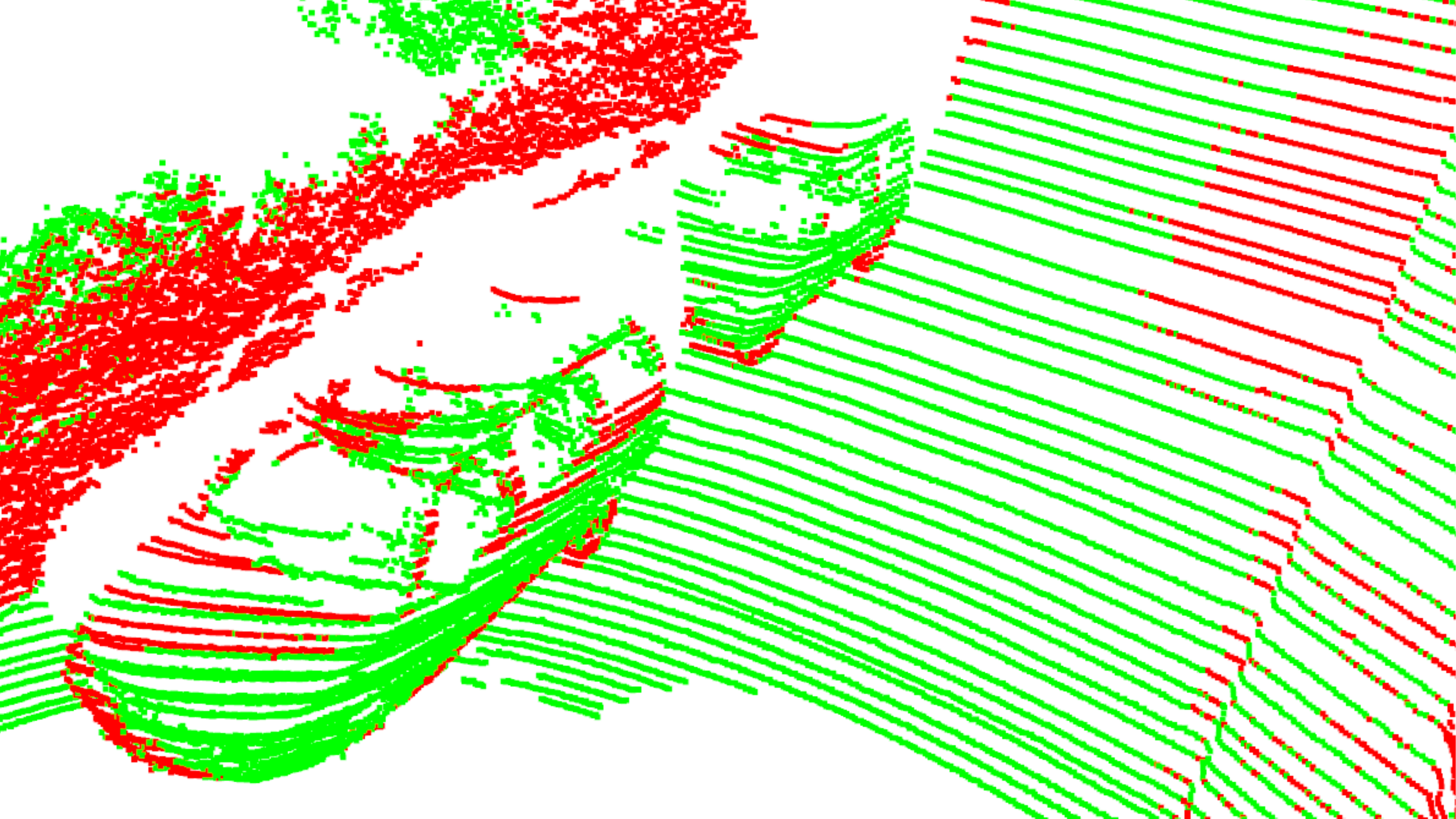}
            \subcaption{}
            \label{fig:pointnet_dif_ali_f1150}
        \end{minipage}
        
        \begin{minipage}[t]{0.25\linewidth}
            \centering
            \includegraphics[keepaspectratio, scale=0.12]{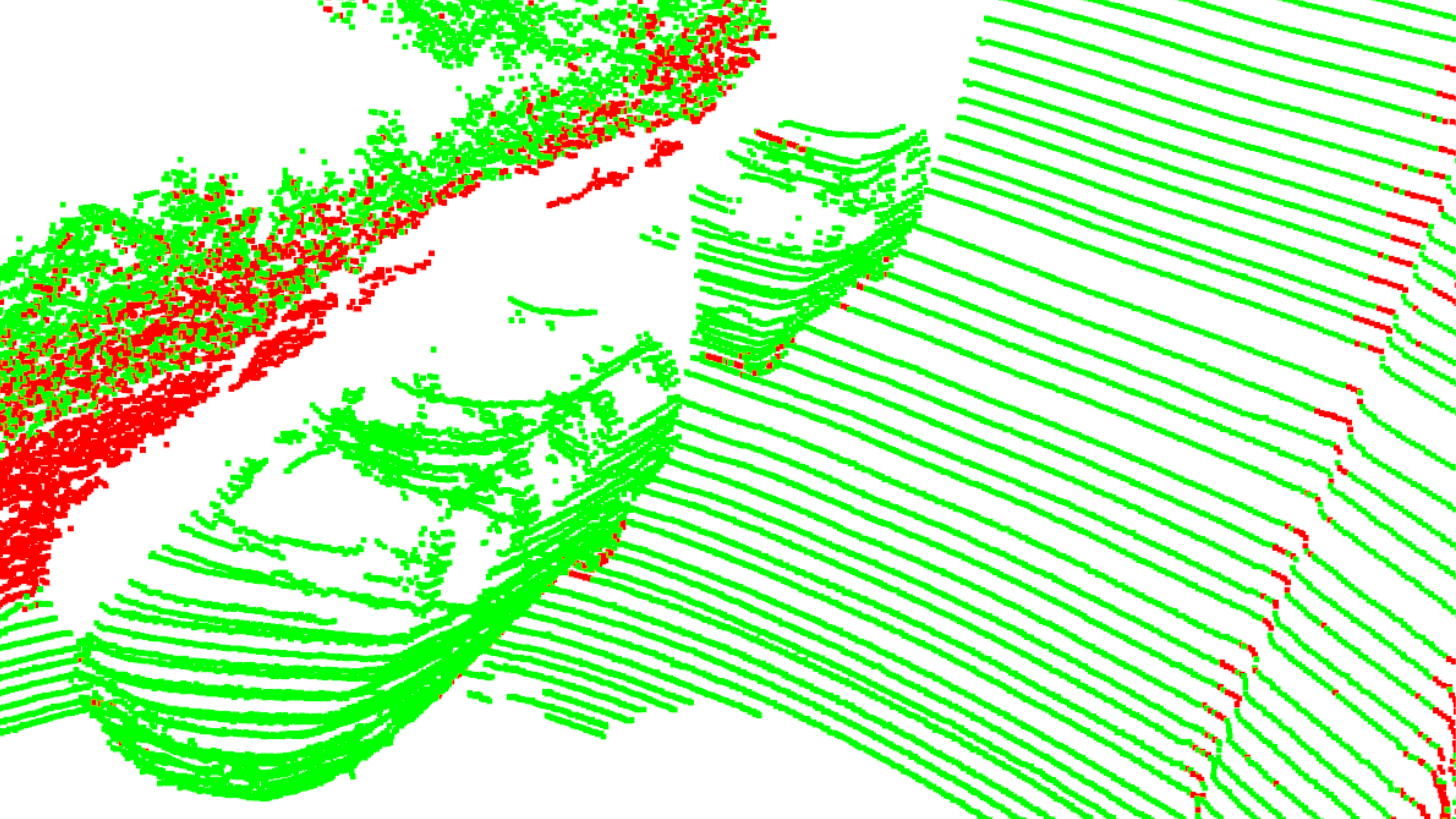}
            \subcaption{}
            \label{fig:pointnet_dif_pp_f1150}
        \end{minipage} 
        
        \begin{minipage}[t]{0.25\linewidth}
            \centering
            \includegraphics[keepaspectratio, scale=0.12]{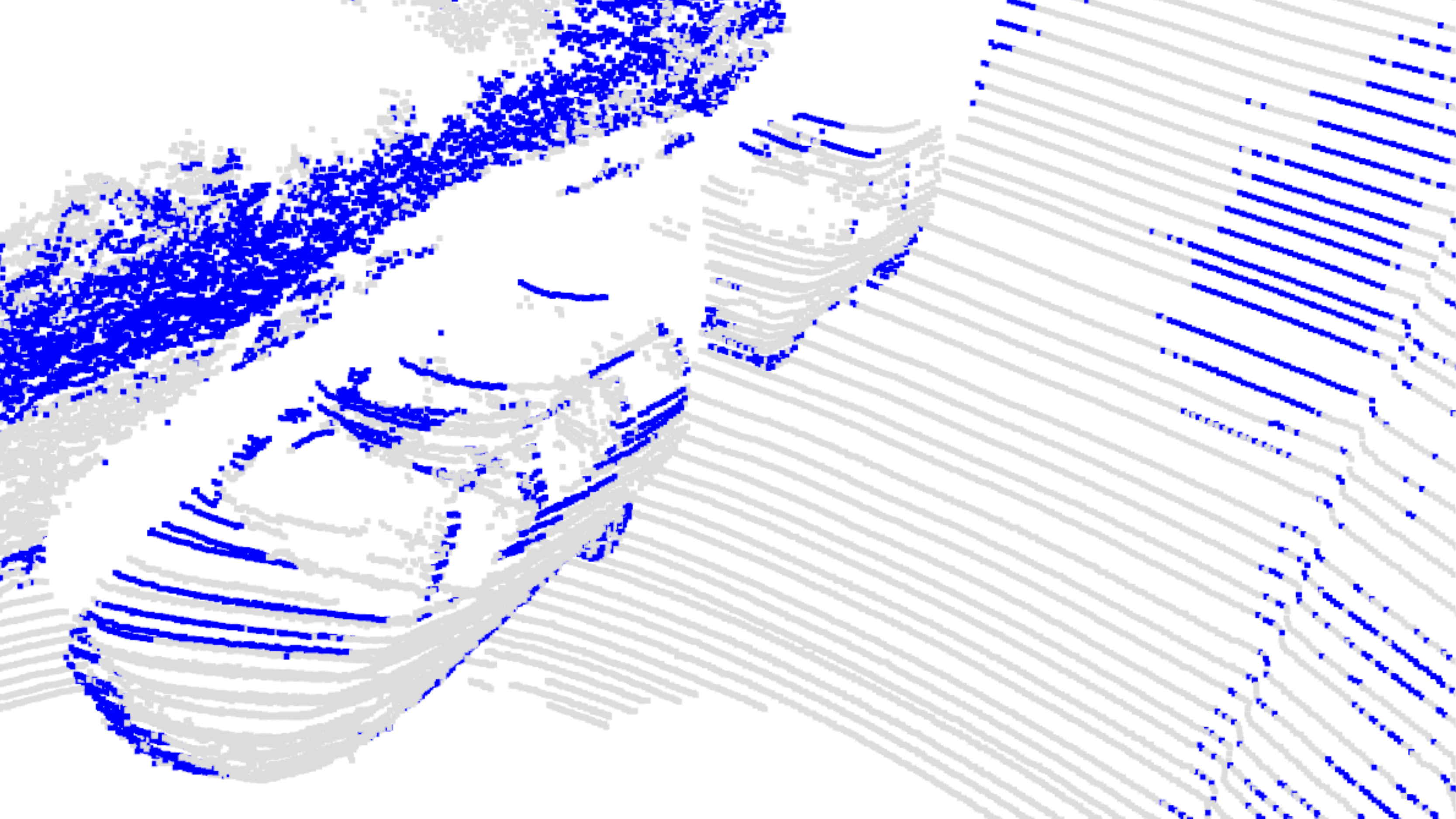}
            \subcaption{}
            \label{fig:pointnet_dif_f1150}
        \end{minipage}
        
        \begin{minipage}[t]{0.25\linewidth}
            \centering
            \includegraphics[keepaspectratio, scale=0.12]{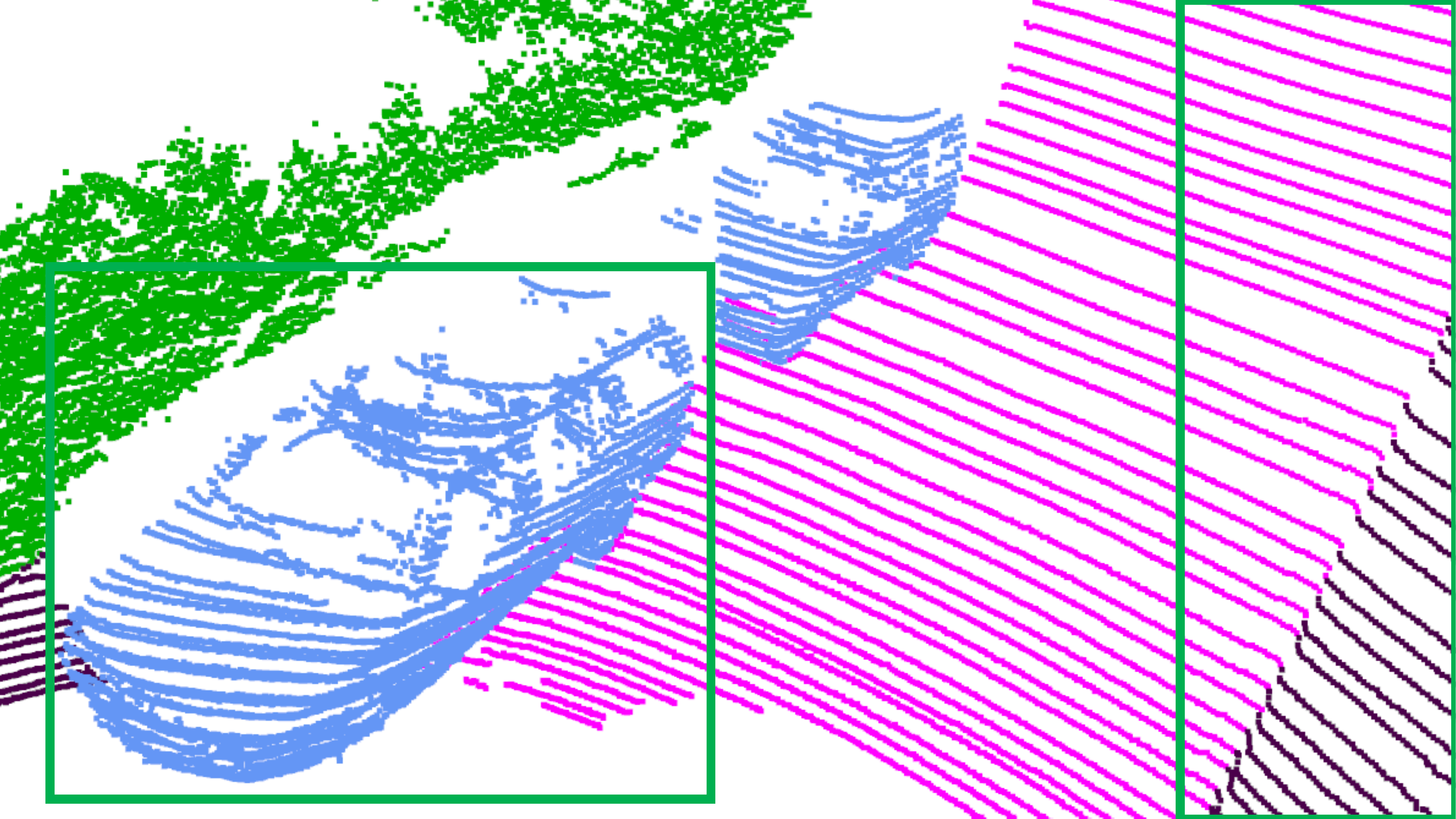}
            \subcaption{}
            \label{fig:out_pointnet}
        \end{minipage}
    \end{tabular}
    \caption{Qualitative evaluation of P$^2$Net for PointNet. (\protect\subref{fig:pointnet_f1148}), (\protect\subref{fig:pointnet_f1149}), and (\protect\subref{fig:pointnet_ali_f1150}) are the sigle-frame prediction results for frames $t-2$, $t-1$, and $t$.  (\protect\subref{fig:pointnet_pp_f1150}) is the refined result by P$^2$Net for frame $t$. (\protect\subref{fig:out_pointnet}) is GT labels of frame $t$. Colors in (\protect\subref{fig:pointnet_f1148})-(\protect\subref{fig:pointnet_pp_f1150}) and (\protect\subref{fig:out_pointnet}) indicate the semantic labels and conform with classes in Tab. \ref{tab:miou}. (\protect\subref{fig:pointnet_dif_ali_f1150}) and (\protect\subref{fig:pointnet_dif_pp_f1150}) show the difference between GT and the predictions w/o and w/ P$^2$Net. {\color{green}Green} color indicates correct predictions while {\color{red}red} color indicates wrongly predicted points. We can see much more {\color{green}green} points, especially the car and road part, in (\protect\subref{fig:pointnet_dif_pp_f1150}) which means great improvement of the accuracy after the refinement by P$^2$Net. Moreover, {\color{blue}blue} colors in  (\protect\subref{fig:pointnet_dif_f1150}) show the points whose labels are refined by P$^2$Net.  }
    \label{fig:refine_pointnet}
\end{figure*}

\begin{figure*}[h!]
    \hspace{-0.5cm}
    \begin{tabular}{cc}
        \begin{minipage}[t]{0.25\linewidth}
            \centering
            \includegraphics[keepaspectratio, scale=0.12]{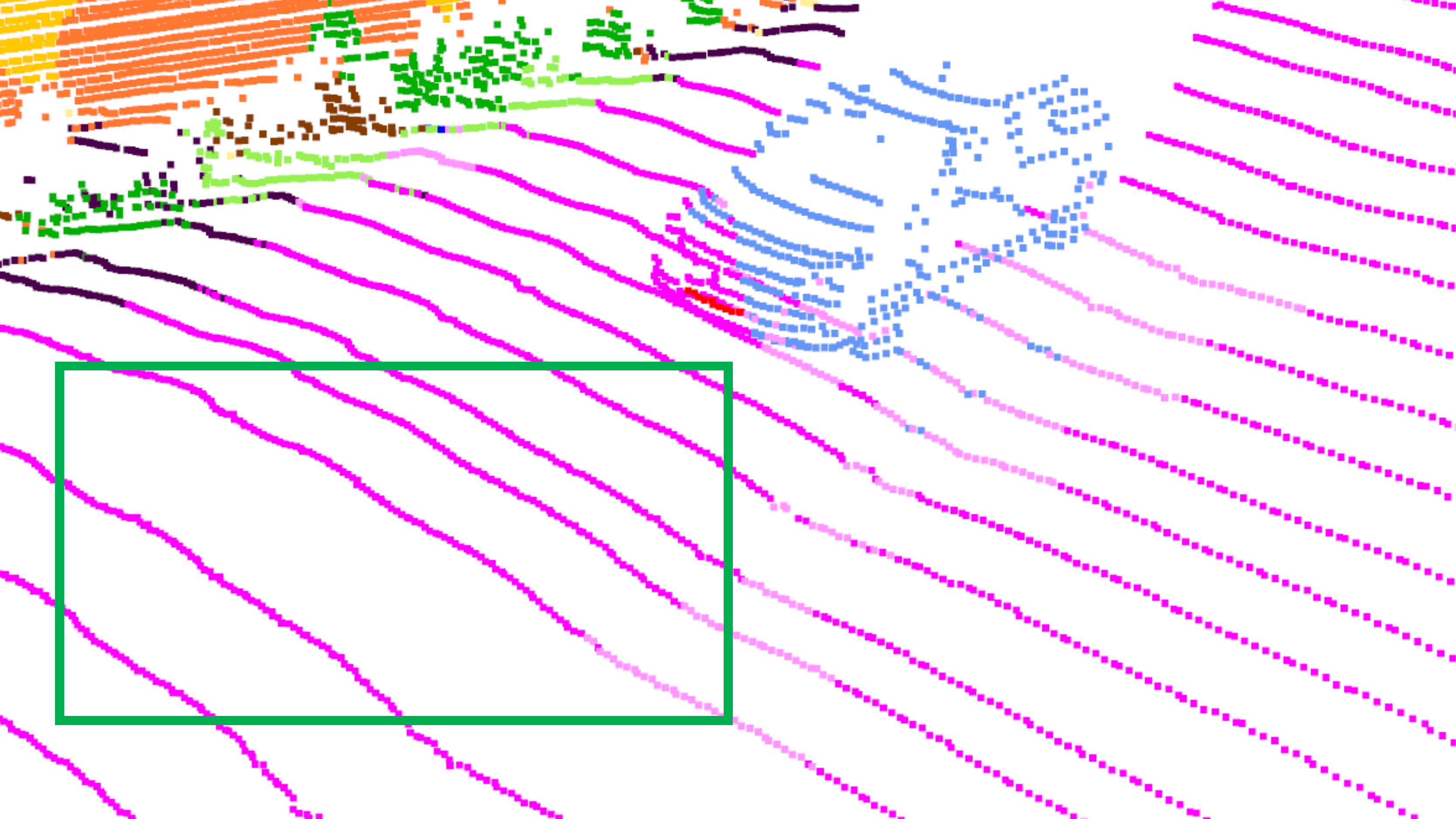}
            \subcaption{}
            \label{fig:pointnet2_f128}
        \end{minipage}
        
        \begin{minipage}[t]{0.25\linewidth}
            \centering
            \includegraphics[keepaspectratio, scale=0.12]{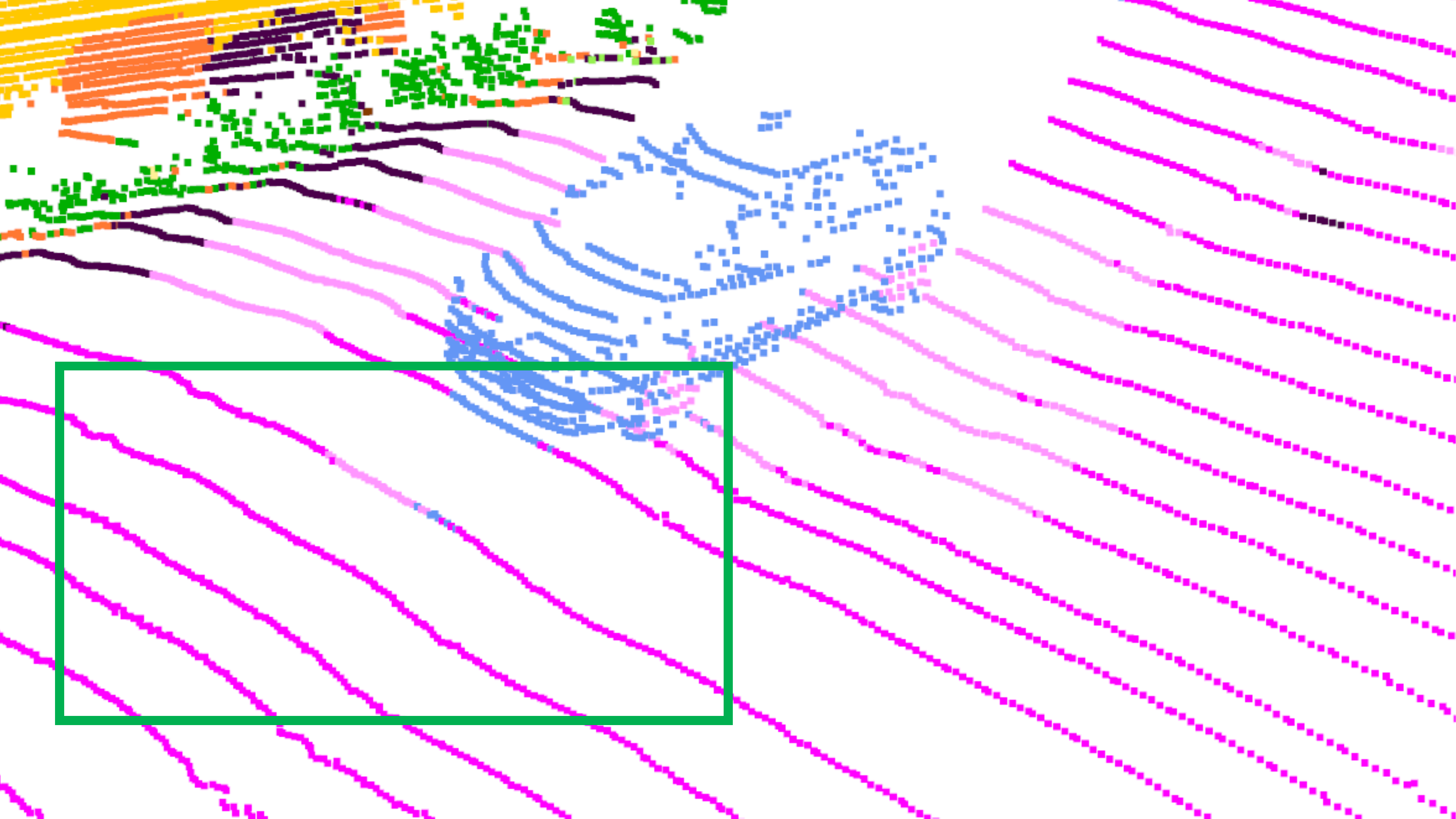}
            \subcaption{}
            \label{fig:pointnet2_f129}
        \end{minipage} 
        
        \begin{minipage}[t]{0.25\linewidth}
            \centering
            \includegraphics[keepaspectratio, scale=0.12]{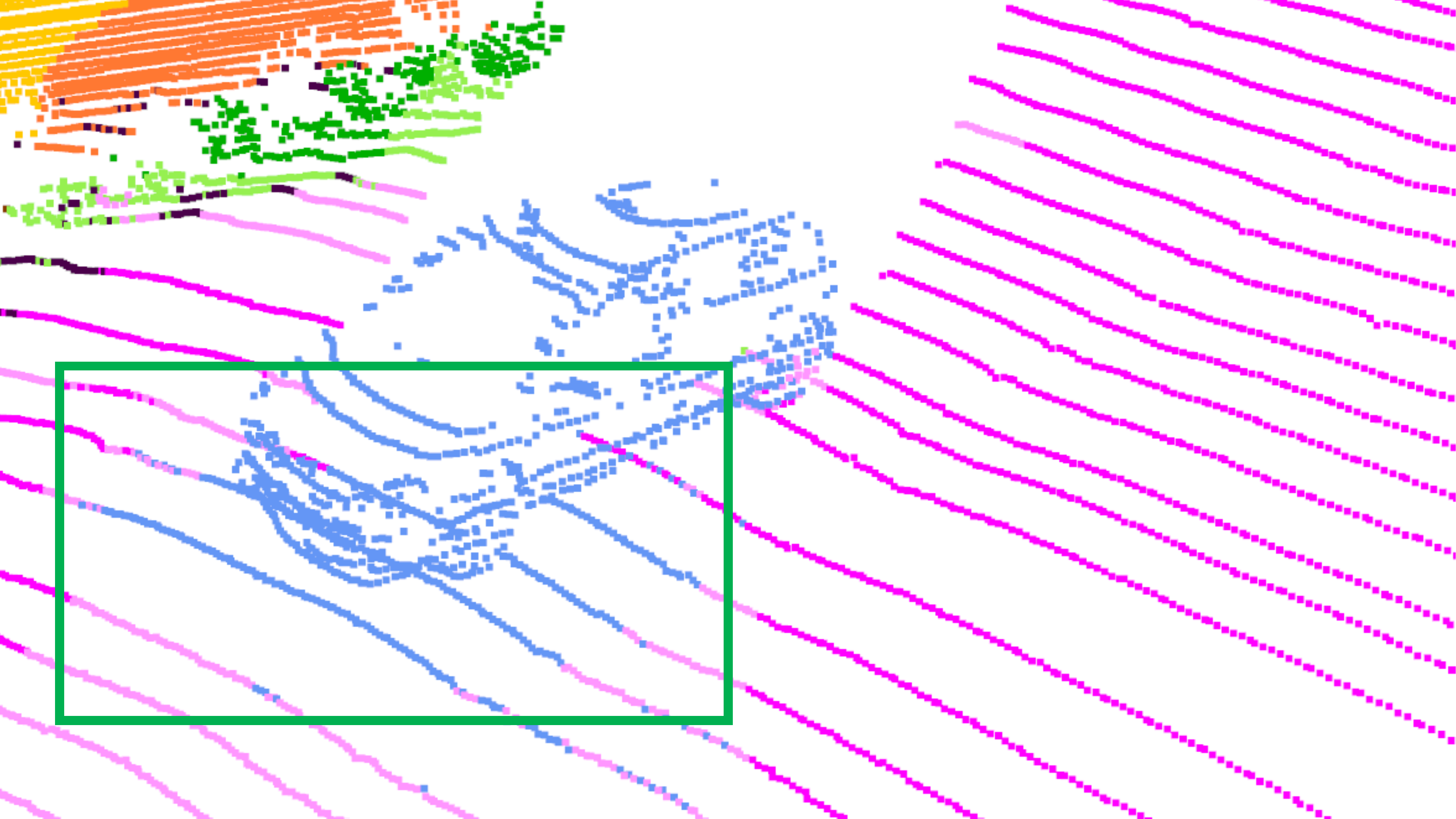}
            \subcaption{}
            \label{fig:pointnet2_ali_f130}
        \end{minipage}
        
        \begin{minipage}[t]{0.25\linewidth}
            \centering
            \includegraphics[keepaspectratio, scale=0.12]{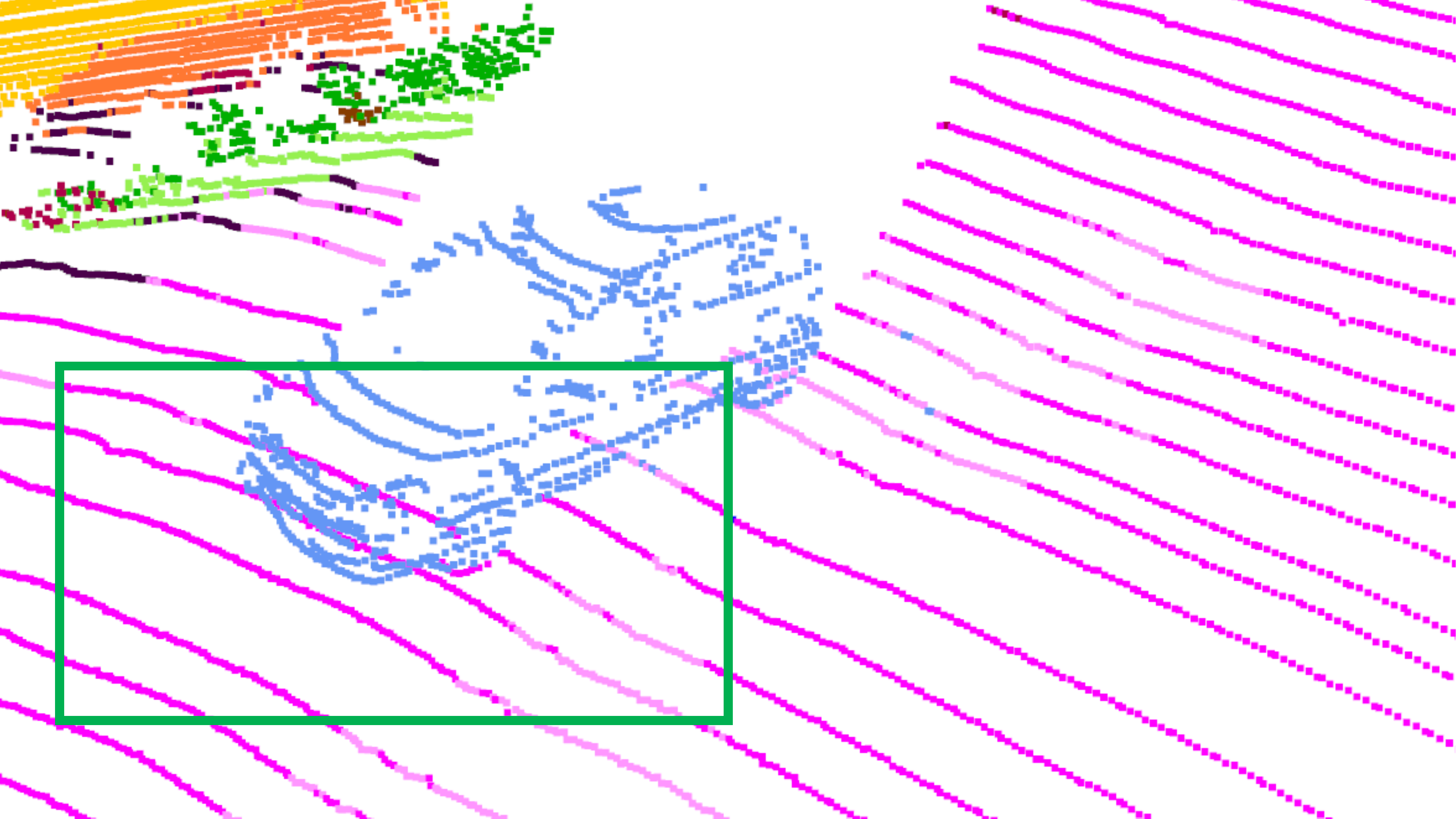}
            \subcaption{}
            \label{fig:pointnet2_pp_f130}
        \end{minipage} \\
        \\
        \vspace{-0.2cm}\begin{minipage}[t]{0.25\linewidth}
            \centering
            \includegraphics[keepaspectratio, scale=0.12]{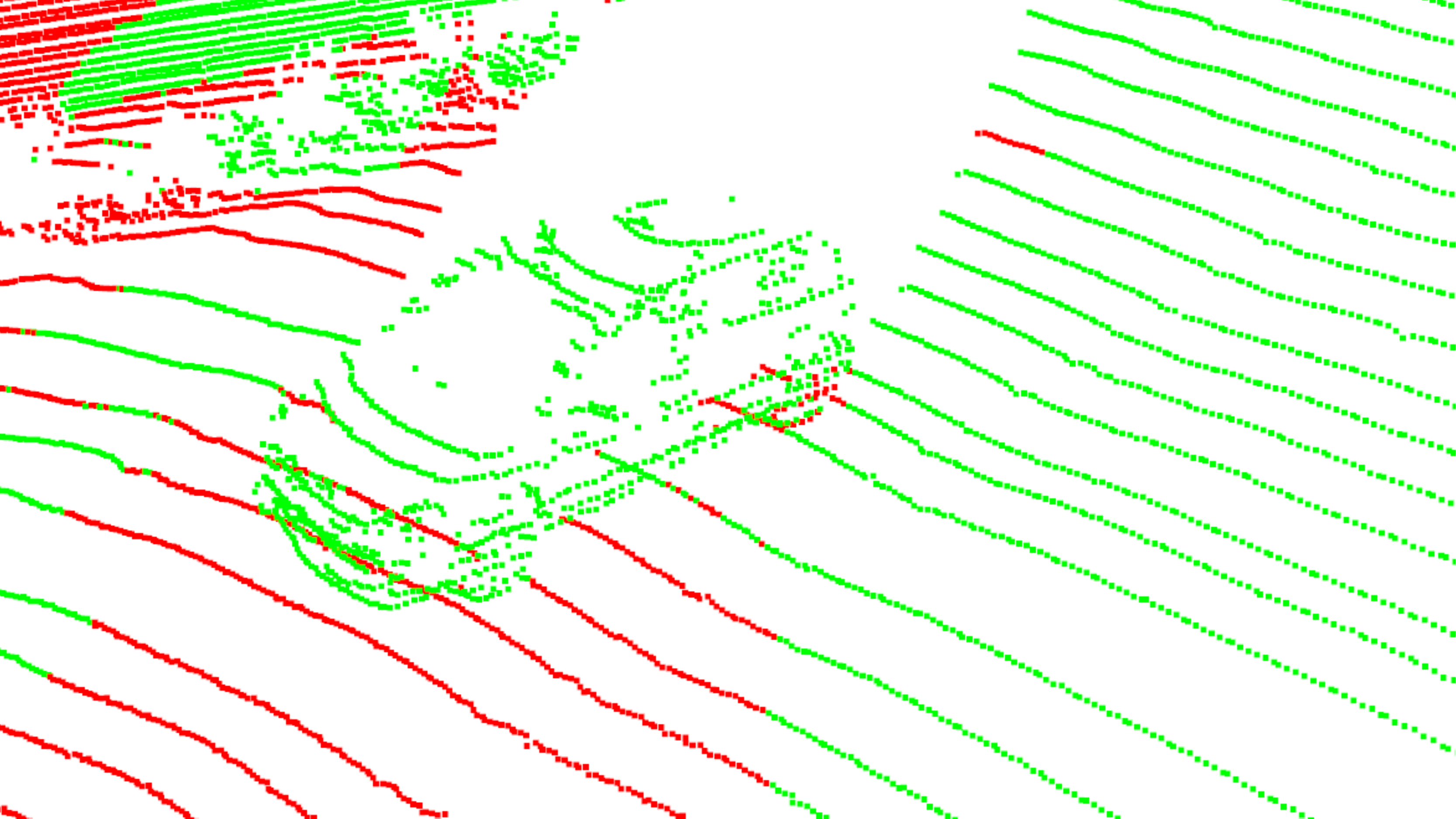}
            \subcaption{}
            \label{fig:pointnet2_dif_ali_f130}
        \end{minipage}
        
        \begin{minipage}[t]{0.25\linewidth}
            \centering
            \includegraphics[keepaspectratio, scale=0.12]{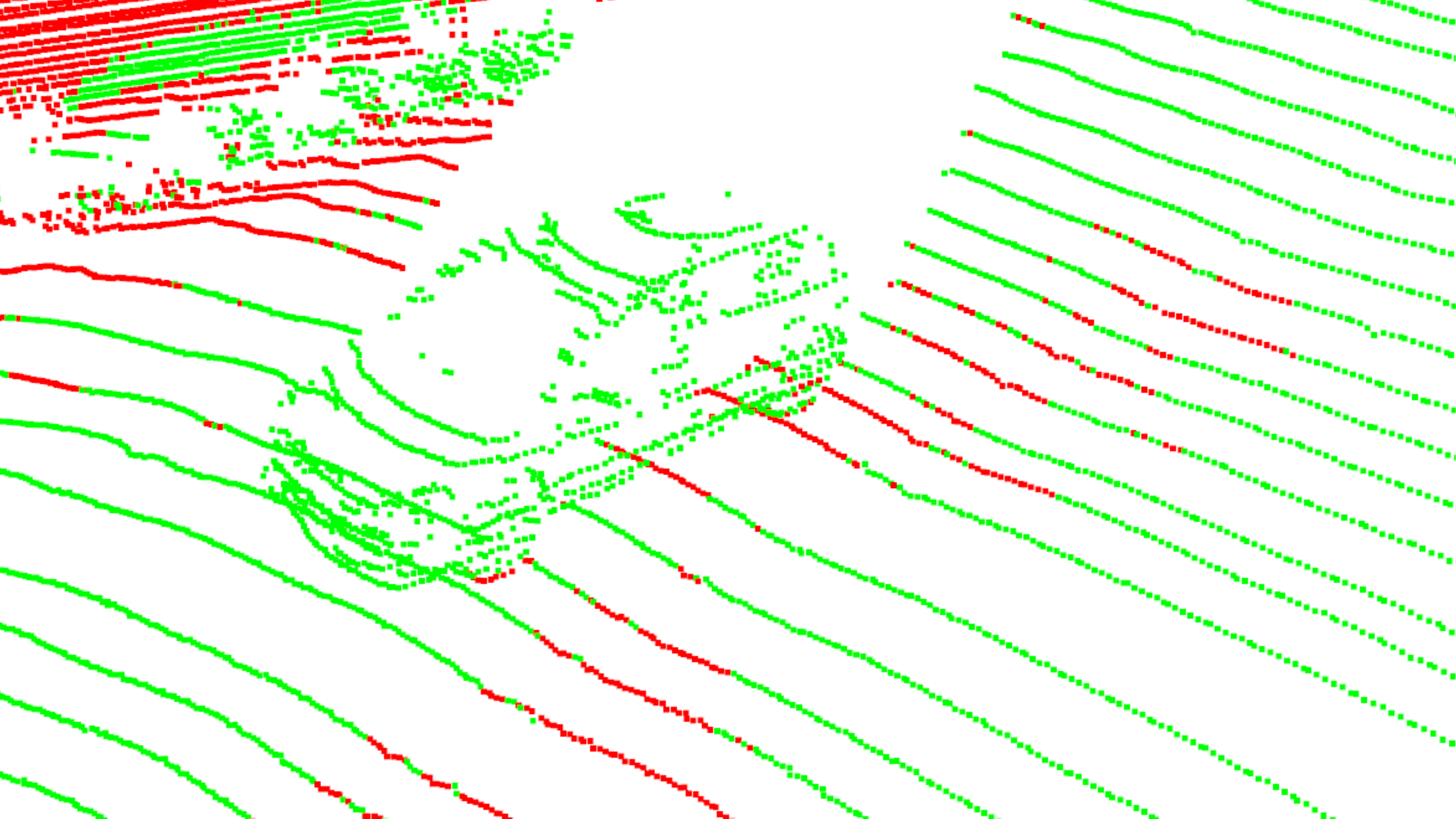}
            \subcaption{}
            \label{fig:pointnet2_dif_pp_f130}
        \end{minipage} 
        
        \begin{minipage}[t]{0.25\linewidth}
            \centering
            \includegraphics[keepaspectratio, scale=0.12]{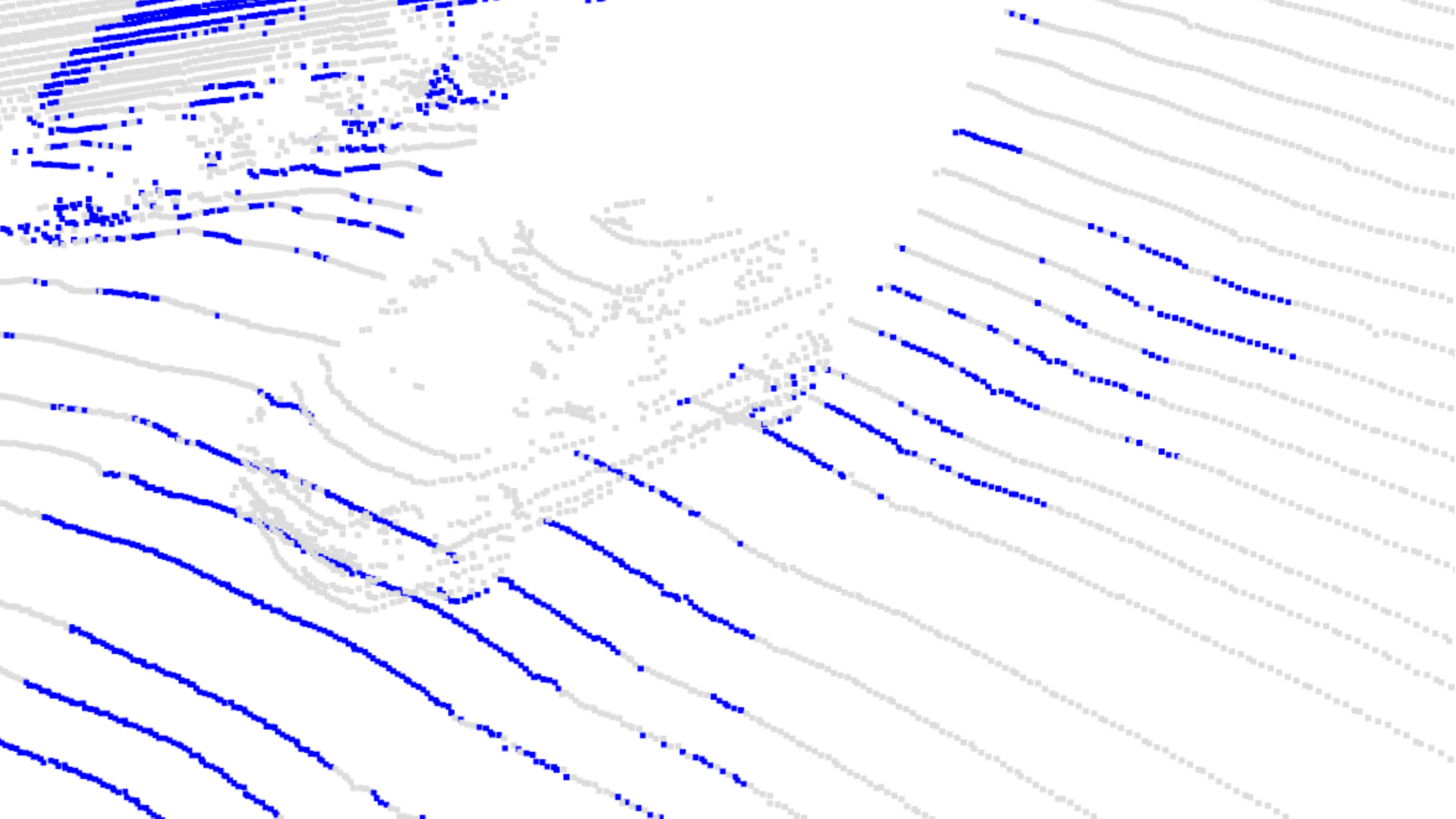}
            \subcaption{}
            \label{fig:pointnet2_dif_f130}
        \end{minipage}
        
        \begin{minipage}[t]{0.25\linewidth}
            \centering
            \includegraphics[keepaspectratio, scale=0.12]{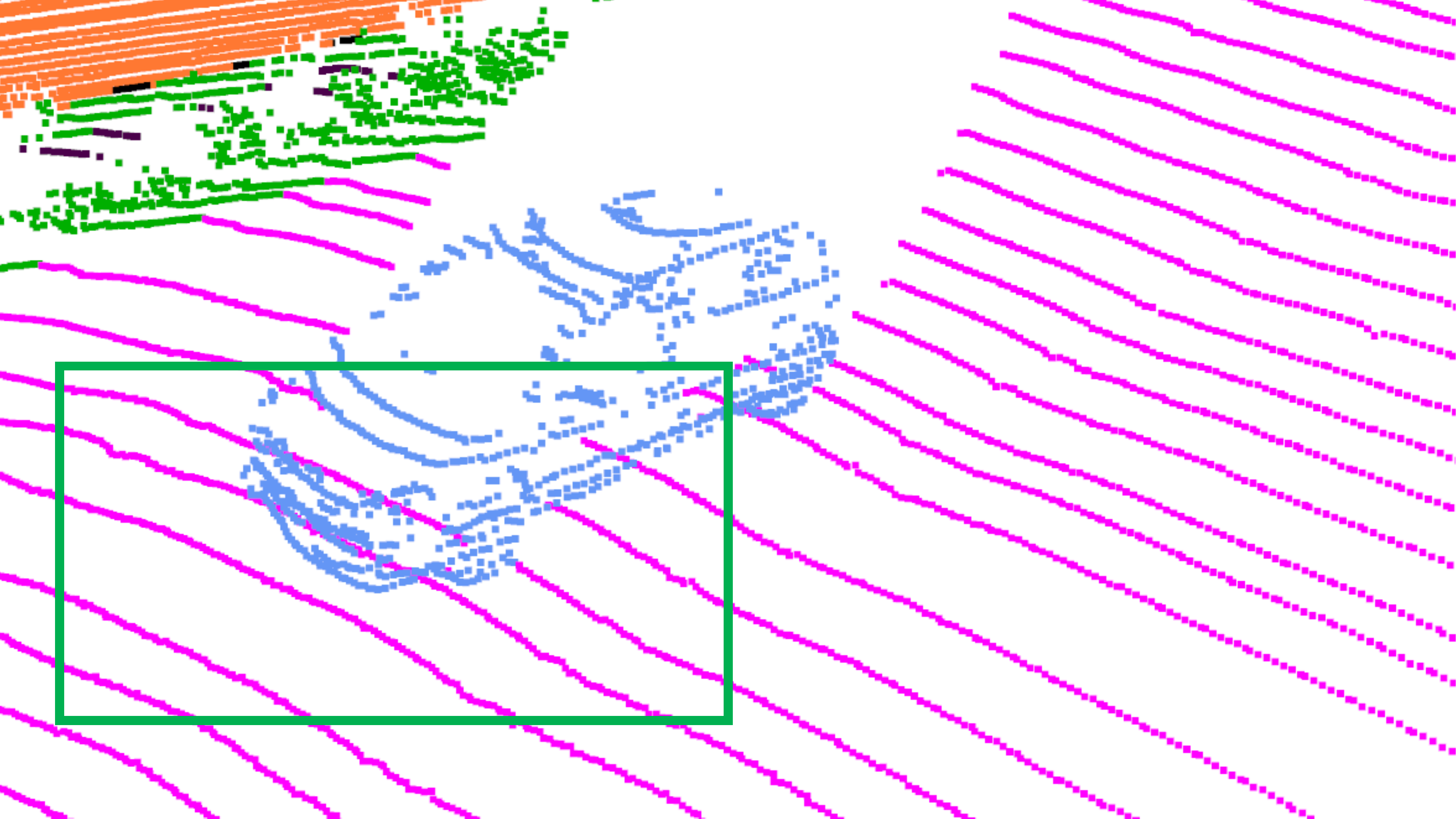}
            \subcaption{}
            \label{fig:out_pointnet2}
        \end{minipage}
    \end{tabular}
        
    \caption{Qualitative evaluation of P$^2$Net for PointNet++. The order and the colors of the panels are the same as those in Fig. \protect\ref{fig:refine_pointnet} other than that predicted results are based on PointNet++. Interestingly, as bounded by the green box in this example scene, we find that points under the car are wrongly predicted in (\protect\subref{fig:pointnet2_ali_f130}) and corrected after the refinement by P$^2$Net while they are correctly predicted in (\protect\subref{fig:pointnet2_f128}) and (\protect\subref{fig:pointnet2_f129}). On the other hand, there also exist points that are wrongly modified by P$^2$Net from {\textcolor[rgb]{1.0, 0.0, 1.0}{\textbf{road}}} to {\textcolor[rgb]{1.0, 0.59, 1.0}{\textbf{parking}}} as shown in (\protect\subref{fig:pointnet2_dif_pp_f130}).}
    \vspace{-0.3cm}
    \label{fig:refine_pointnet2}
\end{figure*}

\subsection{Baseline Setup and Implementation details}

We select PointNet \cite{qi2017pointnet} and PointNet++ \cite{qi2017pointnetplusplus} as baseline methods for the following reasons: 1) They are representative methods that can directly consume 3D point clouds to extract global and local features. 2) The baseline pre-trained models in \cite{behley2019iccv} are not released yet, so we need to train the models. These two networks are easy to setup. 3) Since the purpose of $P^2$Net is to refine the results instead of performing segmentation directly, the illustration of the relative improvement after refining to validate the feasibility is more important rather than the absolute accuracy.

Hyperparameters used for training baseline networks and P$^2$Net are listed in Tab. \ref{tab:train}. The numbers of points within a single scan for training PointNet \cite{qi2017pointnet} and PointNet++ are set to the same numbers with that in SemanticKITTI for memory concerns. As for the $P^2$Net, we increase the number due to the reduction of model parameters.
Adam \cite{kingma2014adam} is used as the optimizer. All models are trained on two GTX 2080Ti
GPUs. We utilize FAISS \cite{JDH17}, an open-source library for efficient similarity search with GPUs developed by Facebook, for the nearest neighbor search of points from different frames.

\subsection{Segmentation Results}
\subsubsection{Quantitative evaluation}

We utilize the mean intersection over-union (mIoU) as the evaluation metric which conforms with that in \cite{behley2019iccv}. Specifically, mIoU is defined as follows: 
\begin{equation}
\label{eq:miou}
  \frac{1}{C}\sum_{c=1}^{C}\frac{\mathrm{TP}_{c}}{\mathrm{TP}_{c}+\mathrm{FP}_{c}+\mathrm{FN}_{c}},
\end{equation}
where $C=19$ is the number of valid classes in SemanticKITTI dataset. $\mathrm{TP}_{c}$, $\mathrm{FP}_{c}$ and $\mathrm{FN}_{c}$ indicates the number of points with predicted True
Positive, False Positive and False Negative labels for class $c$, respectively.

Detailed quantitative numbers are listed in Tab. \ref{tab:miou}. The overall mIoUs of PointNet and PointNet++ are lower than the baselines in \cite{behley2019iccv}. This may be caused by the lack of training data compared with \cite{behley2019iccv} as described in \ref{sec:dataset}. However, we can still evaluate the proposed method by relatively comparing results with and without the refinement by P$^2$Net. From Tab. \ref{tab:miou}, we can see the accuracy improvement with P$^2$Net from $10.5\%$ to $11.7\%$ for PointNet \cite{qi2017pointnet} and from $10.8\%$ to $15.9\%$ for PointNet and PointNet++ respectively.

Basically, P$^2$Net is trained separately for PoinNet and PointNet++. Interestingly, we find that the performance can be improved even when applying a P$^2$Net model trained with PointNet++ to refine the PointNet and vice versa. Nevertheless, the model trained with the corresponding networks gives higher improvement.

\subsubsection{Qualitative evaluation}
In order to show the refinement effects of P$^2$Net more intuitively, we pick up example visualized results and show them in Fig. \ref{fig:refine_pointnet} and \ref{fig:refine_pointnet2}.
Direct predicted results by PointNet and PointNet++ as well as refined results for them are visualized. Besides, we also visualize the refined points by P$^2$Net for a clearer view. The effectiveness of P$^2$Net can be qualitatively validated by comparing panels (e) and (f) in each figure. Furthermore, by checking the points bounded with green boxes in Fig. \ref{fig:refine_pointnet} and \ref{fig:refine_pointnet2}, we can find that P$^2$Net learned to refine labels of points that are difficult to predict due to occlusions or sparseness in the current frame, with predicted labels of nearest neighboring points which are easy to observe in previous frames.
\subsubsection{Time consumption}
The time consumption details to get the refinement are listed in Tab. \ref{tab:time}.  Since P$^2$Net is a purely MLP based structure, it only takes $1.27$ms to refine the predictions with results from two previous frames. The nearest-neighbor search process costs about $1000$ ms, which is much more than the prediction time of PointNet but approximates $50\%$ of the prediction time for PointNet++. A full-frame search is performed in this work and the search process is not optimized yet. There is much room to increase the speed of the nearest search process by reducing the search range. Moreover, parallel processing for independent points can also be applied to increase effectiveness.

\begin{table}[h!]
    \centering
    \renewcommand{\arraystretch}{1.4}
    \begin{tabular}{l|c}
        \hline
        approach & consumption time [ms/frame] \\
        \hline
        PointNet &  34.57\\
        PointNet++ & 2021.24\\
        Nearest neighbor search & 1039.79 \\
        $P^2$Net & 1.27\\
        \hline
    \end{tabular}
    \caption{Time consumption in each process.}
    \label{tab:time}
\end{table}

    \vspace{-0.15cm}
\section{CONCLUSION}
In this paper, we proposed P$^2$Net, which learns to refine the classification predictions based on the consistency of points from multiple sequential frames. Due to the input and structure of the proposed method, it is possible to easily extend the method to any existing segmentation for refining the prediction results. Comprehensive experimental results on the SemanticKITTI dataset demonstrate the feasibility and effectiveness of P$^2$Net. Improvements in the mIoU accuracy of the segmentation results after the refinement by P$^2$Net were confirmed qualitatively. Some visualization examples, labels of the points that are difficult to predict due to occlusion and corrected with information from other frames, also show the qualitative improvements brought by $P^2$Net from the comparison of segmentation results before and after the refinement. 
Our approach succeeds to correct prediction labels and improves the mIoU accuracy with the proposed features combination method. As future work, it is interesting to investigate the relationship between performance improvement and the involved number of frames.

\addtolength{\textheight}{-12cm}   %

\bibliographystyle{IEEEtran}
\bibliography{ref}

\end{document}